\pdfoutput=1

\documentclass[11pt]{article}

\usepackage{ACL2023}

\usepackage{times}
\usepackage{latexsym}
\usepackage{color}
\usepackage{tabularx}
\usepackage[T1]{fontenc}

\usepackage[utf8]{inputenc}

\usepackage{microtype}
\usepackage{CJKutf8}

\usepackage{graphicx} 
\usepackage{epstopdf}
\usepackage{inconsolata}
\usepackage{booktabs} 
\usepackage{multirow}
\usepackage{makecell}
\usepackage{tablefootnote}
\usepackage[para,online,flushleft]{threeparttable}

\definecolor{hzw}{RGB}{223, 97, 76}

\title{TeCS: A Dataset and Benchmark for Tense Consistency \\of Machine Translation}

\author{Yiming Ai,   Zhiwei He,   Kai Yu,   and Rui Wang\thanks{\ \ Corresponding author} \\
  Shanghai Jiao Tong University \\
  \texttt{$\{$aiyiming, zwhe.cs, kai.yu, wangrui12$\}$@sjtu.edu.cn}\\
  }

\begin{document}
\maketitle
\begin{abstract}
Tense inconsistency frequently occurs in machine translation. However, there are few criteria to assess the model's mastery of tense prediction from a linguistic perspective. In this paper, we present a parallel tense test set, containing French-English 552 utterances\footnote{ The following updates will be shown at:

\url{https://github.com/rutilel/TeCS-A-Dataset-and-Benchmark-for-Tense-Consistency}. }. We also introduce a corresponding benchmark, tense prediction accuracy. With the tense test set and the benchmark, researchers are able to measure the tense consistency performance of machine translation systems for the first time. 
\end{abstract}

\begin{CJK*}{UTF8}{gbsn}
\section{Introduction}\label{sec1}

Translation tools are often found in a variety of social situations to enable cross-linguistic communication. Tenses are used to express time relative to the moment of speaking. Human translators frequently pay close attention to tense correspondence \citep{gagne2020english}. Similarly, machine translation (MT) systems are supposed to maintain temporal consistency between the original text and the predicted text to avoid misunderstandings by users. However, accurately keeping the tense consistency is undoubtedly difficult. Taking French-English (one of the most classic language pairs for MT) as an example in Table \ref{table1}, the original text is in \textit{plus-que-parfait de l’indicatif} of French, corresponding to the \textit{past perfect} tense in English, while the English prediction provided by Google Translator is in the \textit{past simple} tense.

\begin{table}[!htbp]
\centering
\begin{tabular}{p{0.55\linewidth} | p{0.32\linewidth}}
\hline
\textbf{Sentence} & \textbf{Tense}\\
\hline
FR: Mais on les avait votés lors de la dernière période de session. & \textit{Plus-que-parfait}  \\
\hline
EN:  But we voted on them during the last part-session. &  \textit{Past simple} \\
\hline
Correction: But we had voted on them during the last part-session. & \textit{Past perfect}  \\
\hline
\end{tabular}
\caption{\label{table1}
An example of  tense corrspondence in machine translation
}
\end{table}

In fact, this is not an isolated case. You can also find several examples in Appendix \ref{sec:appendix}. Besides. the translation mechanics may not the only reason leading to tense inconsistency. The corpora matter as well. For example, we have extracted 20,000 pairs English-French parellel sentences from the widely used dataset Europarl \citep{koehn2005europarl}, and we have observed all groups of parallel utterances where the original French texts are in the \textit{plus-que-parfait de l’indicatif} tense, examining the tenses of their English counterparts. As a sentence may include several tenses, there are 195 occurences of \textit{plus-que-parfait} tense in total. Among them, only $35.28\%$ English sentences are in the correct \textit{past perfect} tense, as shown in Table \ref{table2}. Although, compared to other tense correspondences, the pair of \textit{plus-que-parfait} and \textit{past-perfect} is prone to error in datasets and there are only 0.94$\%$ of sentences in Europarl are in plus-que-parfait, we cannot easily ignore this issue. Like Europarl, tense correspondences are generally credible but unreasonable for certain tenses in several common datasets.

\begin{table}[!htbp]
\centering
\begin{tabular}{lr}
\hline
\textbf{Tense of Counterpart} & \textbf{Proportion}\\
\hline
Past perfect (correct) & $35.28\%$ \\
Past simple & $54.46\%$ \\
Present perfect & $8.21\%$ \\ 
Present & $2.05\%$ \\ 
\hline
\end{tabular}
\caption{Preliminary statistics of  translation tense }
\label{table2}
\end{table}

\begin{table*}[!htbp]\small
\centering
\begin{tabular}{p{0.25\linewidth} | p{0.25\linewidth} | p{0.10\linewidth} | p{0.30\linewidth}}
\hline
\textbf{French Tenses} & \textbf{English Tense} & \textbf{Format} & \textbf{Example}\\
\hline
Imparfait, Passé composé, Passé simple, Passé récent & Past simple / progressive & \textit{Past} & That \textbf{was} the third point.\\
\hline
Présent, Future proche & Present simple / progressive & \textit{Present} & The world \textbf{is changing}.\\
\hline
Future simple, Future proche & Future simple / progressive & \textit{Future} & I \textbf{will communicate} it to the Council.\\
\hline
Plus-que-parfait & Past perfect & \textit{PasPerfect} & His participation \textbf{had been notified}.\\
\hline
Passé composé & Present perfect & \textit{Preperfect} & This phenomenon \textbf{has become} a major threat.\\
\hline
Future antérieur & Future perfect & \textit{Futperfect} & We \textbf{will have finished} it at that time.\\ 
\hline
Subjonctif, Conditionnel & including Modal verbs & \textit{Modal} & We \textbf{should be} less rigid.\\
\hline
\end{tabular}
\caption{\label{table4}
French-English tense pairs, annotation format of English tenses and corresponding example sentences
\textit{(Where the modal verb contains can, may, shall, must, could, might, should and would.)}
}
\end{table*}

In addition to the train set, the difficulty of remaining tense consistency also stems from the lack of metrics on measuring the model's mastery of tense information. The research of \citet{marie2021scientific} shows that $98.8\%$ of *ACL papers\footnote{The papers only includes *ACL main conferences, namely ACL, NAACL, EACL, EMNLP, CoNLL, and AACL.} in the field of MT from 2010 to 2020 used BLEU \cite{papineni-etal-2002-bleu} scores to evaluate their models. However, the reliability of BLEU has been questioned in the era of neural machine translation (NMT) as its variants only assess surface linguistic features \citep{shterionov2018human}, and many studies have shown that BLEU has difficulty in portraying the degree of semantic information mastered by the model, i.e. its score does not necessarily improve when more semantic information is mastered \citep{mathur2020tangled,he2023exploring}, not to mention specific tense information. We have also applied BLEU to measure various baselines on our tense test set in Section \ref{lab:exp}, and the results explicitly support the above statement. In addition, reviewing the evaluation criteria related to MT tasks over the past ten years, we are surprised to find that there are no criteria to assess the model's mastery of tense prediction from a linguistic perspective.

Therefore, our paper is devoted to the study of 
NMT based on semantic understanding in terms of tense. We construct a tense parallel corpus test set consisting of 552 pairs of tense-rich, error-prone parallel utterances for NMT systems, and then propose a new task for evaluating the effectiveness of model translations from the perspective of tense consistency. This paper makes three contributions: (1) the presentation of the construction of the tense test set, including its tense labels; (2) the proposal of a feasible and reproducible benchmark for measuring the tense consistency performance of NMT systems; and (3) the various experiments for different baselines with the above test set and corresponding benchmark.

\section{Annotation Rules and Tools}
\label{sec2}
 As the first work of the MT tense study, we choose English-French, one of the most classic language pairs of MT, to construct the dataset\footnote{Please refer to the ~\nameref{sec:limit} for more details.}.
 
 TENSE, the dominant topic of our research, is a combination of tense and aspect. In the modern grammar system of English, ``a tense system is a system associated with the verb where the basic contrasts in meaning have to do with the location in time of the situation, or the part of it under consideration'' \citep{huddleston2021student}. The modern grammatical system divides tense into present and preterit based on the inflections added to the end of verbs, and the aspect into perfective and progressive on the state where an action is \citep{kamp1991tense}. While this tense classification system is too crude for daily life, we therefore apply the following classification methods. On the one hand, we classify the tenses according to the macro-temporal interval of the action into three major time intervals, namely present, past and future tenses; on the other hand, we classify the tenses according to the state of the action into general, progressive and perfect aspects. Hence, 9 kinds of tenses are born through combining the three tenses and the three aspects. 

French and English belong to the same Indo-European language family and share many similarities in various respects. The main difference is that in French there is another grammatical point called \textit{mode}, part of which is like the \textit{aspect} in English. In terms of tenses, we will generally discuss the tenses in the indicative mode of French and will describe the others later in this section. In the following, if there is no mode qualifier before a tense, it is by default in the indicative mode. Careful identification and comparison of the subdivided tenses in the three main tense intervals, English and French, reveals a very similar usage of the tenses, as summarised in Table \ref{table4}. As there is no progressive tense in French, we do not distinguish the progressive tense in English, but rather merge the progressive tense into its corresponding base tense, e.g. the present perfect progressive tense into the category of the present perfect tense.

When discussing tenses from a semantic point of view, the modes also need to be taken into account.  The grammatical correlations between French and English modes are quite complicated. Considering the corresponding grammatical expressions of 2 modes strongly related to tense, \textit{conditionnel} and \textit{subjonctif}, in French rely on the usage of modal verbs, we introduce \textit{modal verbs} to simplify the distinguishment of the modes.


Based on these grammatical rules, we merge the nine common tenses in English into seven categories that correspond reasonably and rigorously to French, namely the 6 tense categories of \textbf{past/present/future + simple/perfect} and statements containing \textit{modal} verbs that correspond to the French \textit{subjonctif} and \textit{conditionnel} tenses. We construct an automatic annotation method based on the spaCy package \citep{Honnibal_spaCy_Industrial-strength_Natural_2020}. First, we label the grammatical components of each word in the sentence based on the spaCy package, and then we define and compare the grammatical structures of the verb phrases with the structures of each tense classification to derive the sentence tense labels. During this process, to simplify the annotation process and better correspond with French \textit{futur proche} tense, we classify the expression ‘\textit{be going to do}’, grammatically in Future tense, into the Present tense, just like expressions ‘\textit{be about to do}’ and ‘\textit{be + verb progressive}’, whose stucture are in \textit{Present} tense but the real meaning is about the close future. Also, a sentence may have several tense structures, in this case, the tense label consists several tenses. For example, the label of the sentence ‘\textit{So it is in that spirit that we have made this change}.’ is ‘\textit{Present+PrePerfect}’. 

\section{Corpus Design and Characteristics}
\subsection{Corpus Design}
\label{sec3.1}

We choose the tense-rich Europarl, namely EuroparlPV, processed by \citet{loaiciga2014english} as the source corpus, for it contains all the sentences with predicate verb structures in the original Europarl dataset \citep{koehn2005europarl}. First, we cleaned the source corpus, including deleting sentences without counterparts, English sentences in the French part and vice versa. After this, we obtain 201,374 tense-rich parallel French-English sentence pairs, namely EuroparlTR. We randomly divided them into a training set, a validation set and a test set in the ratio of 8:1:1, and trained a transformer baseline based on this using fairseq \citep{ott2019fairseq} with a BLEU value of 33.41. Then we compared a total of 20,000 parallel sentences’ triples (\textit{original Europarl French text, original Europarl English text, transformer English prediction}).

In the construction process, with the code mentioned in Section \ref{sec2}, we first automatically annotated the original English text and English prediction in the 20,000 pairs of parallel utterances, given the corresponding tense labels. Then, we filtered \textbf{6,779} parallel French-English sentence triples with different tense labels for English originals and predictions. On the basis of the automatic selection, we manually screened out the representative parallel French-English sentence pairs with a certain degree of translation difficulty and a complex grammatical structure. We also corrected the reference translations that did not justify the tense or semantics. It is worth noting that the author has a level of English and French that meets the C1 standard of The Common European Framework of Reference for Languages (CEFR), representing the ability to express herself effectively and flexibly in English and French in social, academic and work situations. A total of \textbf{570} parallel pairs of statements were selected at this stage. 

\begin{table}[t]
\centering
\begin{tabular}{lrr}
\hline
\textbf{Classfication} & \textbf{Times} & \textbf{Proportion}\\
\hline
Past & 101 & $12.95\%$ \\
Present & 444 & $56.92\%$ \\
Future & 56 & $7.18\%$ \\ 
Past perfect & 22 & $2.82\%$ \\ 
Present perfect & 43 & $5.52\%$ \\ 
Future perfect & 10 & $1.28\%$ \\ 
Modal & 104 & $13.33\%$ \\ 
\hline
\end{tabular}
\caption{Distribution of 780 tense structures in 552 annotated sentences of the corpus}
\label{table5}
\end{table}

Following this, two other reviewers at CEFR C1 level, reviewed the tense test set for semantic and tense correspondence, and the tense labels marked by the automatic annotation code. The tense test set was further refined. The final test set contains \textbf{552} parallel French-English sentence pairs. You can see more details in Appendix \ref{sec:human}.

\begin{table*}
\centering
\begin{tabular}{l cc cc cc c}
\toprule
\multirow{2}{*}{\bf System} & \multicolumn{2}{c}{\bf Tense set} & \multicolumn{2}{c}{\bf Europarl testset} & \multicolumn{2}{c}{\bf WMT15 testset} & \multirowcell{2}{\bf Tense\\\bf Accuracy} \\
\cmidrule(lr){2-3} \cmidrule(lr){4-5} \cmidrule(lr){6-7} 
                            & BLEU & COMET & BLEU & COMET & BLEU & COMET \\
\midrule
Transformer (tense-rich)    & 47.71 & 0.631 & 27.38 & 0.269 & 14.17 & -0.429 & $66.30\%$ \\
Transformer (tense-poor)    & 43.24 & 0.588 & 27.28 & 0.264 & 14.68 & -0.444 & $58.33\%$ \\
LSTM (tense-rich)           & 44.21 & 0.558 & 25.53 & 0.126 & 12.04 & -0.590 & $67.75\%$ \\
LSTM (tense-poor)           & 41.92 & 0.483 & 26.17 & 0.147 & 12.27 & -0.598 & $58.70\%$ \\
CNN (tense-rich)            & 47.10 & 0.567 & 26.83 & 0.147 & 15.30 & -0.512 & $68.48\%$ \\
CNN (tense-poor)            & 43.23 & 0.502 & 26.95 & 0.144 & 14.96 & -0.525 & $57.97\%$ \\
Bi-Transformer (tense-rich) & 47.10 & 0.632 & 28.17 & 0.295 & 14.72 & -0.392 & $64.13\%$ \\
Bi-Transformer (tense-poor) & 43.87 & 0.578 & 28.30 & 0.298 & 14.39 & -0.428 & $55.25\%$ \\

\midrule
Bing Translator             & 61.72 & 0.895 & - & - & - & - & $77.36\%$ \\
DeepL Translator            & 59.50 & 0.904 & - & - & - & - & $79.02\%$ \\
Google Translator           & 57.00 & 0.878 & - & - & - & - & $81.70\%$ \\
\bottomrule
\end{tabular}
\caption{\label{table6} Experimental results of various baselines and common business translators
}
\end{table*}

\subsection{Corpus Characteristics}

In the following paragraphs, we describe the statistical features of our corpus and the elimination of gender coordination influence.  

\textbf{Tense distribution}. The corpus consists of 780 tense structures in 552 sentences, and the distribution of tense classifications is shown in Table \ref{table5}. In the test set, sentences in present tense are the most, corresponding the situation of the reality: we use present tense most frequently and future perfect sense least frequently. 


\textbf{Elimination of gender effect}. Unlike English, gender coordination exists in French. For example, the French sentences `\textit{Nous nous sommes donc \textbf{abstenus}.}' and `\textit{Nous nous sommes donc \textbf{abstenues}}.' both correspond to the English `\textit{We therefore abstained.}'. That is, the MT system's ability to learn gender coordination affects its ability to recognize tense structures, which in consequence affects the maintenance of tense consistency between original French text and predicted English sentence.
Therefore, to better measure the tense-predicting capability of different MT systems, rather than their ability to recognize pronominal gender, we controlled for the gender variable by defaulting all pronouns, which do not indicate explicitly their genders, as masculine. These pronouns consists of 167 \textit{je} (I), 114 \textit{nous} (we, us) and 28 \textit{vous} (you).

\section{Experimental Results}
\label{lab:exp}
To measure the tense consistency performance of different systems, we introduce a benchmark called \textbf{tense (prediction) accuracy}, as shown in Eq. (\ref{equation}).

\begin{equation}
\label{equation}
    {\rm Accuracy} = \frac{N_c}{N_t},
\end{equation}%
where $N_c$ is the number of predicted utterances with the same tense as its reference and $N_t$ is the total number of utterances in the tense set.

To verify the validity of our tense corpus, the following approach was adopted: To begin with, $100,000$ parallel utterance pairs from the EuroparlTR (containing $201,374$ pairs) mentioned in Section \ref{sec3.1} were extracted as the tense-rich train set, and $100,000$ parallel utterance pairs from the EuroparlPV corpus \citep{koehn2005europarl} were extracted as the tense-poor train set. There were no overlapping utterances between the latter and the former. We performed the same preprocessing procedure, including data cleaning, tokenization and BPE coding. We then trained four pairs of French-English NMT systems with different architectures based on fairseq \citep{ott2019fairseq}, where two systems in each pair differed only in the train set. After this, we summarized the scores evaluated by SacreBLEU \citep{post2018call} and COMET~\citep{rei2020comet} and tense prediction accuracies of the eight systems on different test sets. We have applied three types of test sets: our tense set, the Europarl test set and the WMT15 test set. The Europarl test set contains 3,000 parallel utterance pairs drawn from the Europarl corpus, the exact same field of train set, while the WMT15 is a test set for the WMT15 ~\citep{bojar-etal-2015-findings}, deriving from data in the different field of train set. Besides, we also apply our approach to mesure the tense consistency performance of several business translators, including Bing Translator, DeepL Translator and Google Translator. The results are listed in Table \ref{table6}:

1) The BLEU and COMET scores based on the Europarl set and the WMT15 set are quite similar for each system pair, which indicates that the translation capabilities of the two systems are similar in the general evaluation dimension. This suggests that by relying solely on the difference in BLEU scores on traditional test sets, we are unable to measure the tense prediction ability of the systems.

2) However, there are large differences in our tense set. The tense consistency performance of systems trained on the tense-rich train set was significantly better than that of systems trained on the tense-poor train set.  This indicates that our tense set can capture the tense consistency performance. 

3) Further investigation of the BLEU or COMET) scores and tense prediction accuracy for each system reveals their positive correlation for the same architecture, but not across architectures. To measure the tense consistency performance across different architectures, we should focus more on tense accuracy rather than BLEU scores only.

\section{Conclusion}

We presented the French-English parallel tense test set and introduced the corresponding benchmark \textit{tense prediction accuracy}, providing a brand-new approach to measure the tense consistency performance of machine translation systems. This test set firstly focuses on the tense prediction ability, posing a new dimension to improve the MT quality. 

In the future, we will endeavour to generalize the test set to other languages. Considering there are statements like "the use of tense A in language X is equivalent or similar to the use of tense B in English" in grammar books of other languages\citep{durrell2015essential}, even across language families\citep{gadalla2017translating} and human translators also apply such rules\citep{santos2016translation}, we are confident in taking this forward.

\section*{Limitations}
\label{sec:limit}
In this work, we focus on creating the English-French tense corpus. These two languages are among the most frequently and widely used languages in the world. In addition, they have several similarities in tenses, which are pretty helpful for research on tense consistency through machine translation. Thanks to the distinctive tense structures, the study of these two languages makes it possible to examine many common tense issues, but there are also some tense issues in other languages that are not covered by this study. For example, the implicit tense expressions in Chinese are difficult to correspond to the explicit tense expressions in English \citep{jun2020translation}. Hence, our next step will be to extend the tense test set to other language families and even cross-language families to further study tense consistency.
Also, as for future work, we will optimize both the tense annotation method and the tense prediction accuracy calculation. Besides, we did not propose a new method to improve the tense prediction accuracy. To be further, we will endeavour to improve the existing machine translation systems according to tense consistency.

\section*{Acknowledgements}
Yiming, Zhiwei, and Rui are with MT-Lab, Department of Computer Science and Engineering,
School of Electronic Information and Electrical Engineering, and also with the MoE Key Lab of Artificial Intelligence, AI Institute, Shanghai Jiao Tong
University, Shanghai 200204, China. Rui is supported by the General Program of National Natural Science Foundation of China (6217020129), Shanghai
Pujiang Program (21PJ1406800), Shanghai
Municipal Science and Technology Major Project
(2021SHZDZX0102), Beijing Academy of Artificial Intelligence (BAAI) (No. 4), CCF-Baidu Open Fund (No. CCF-BAIDU OF2022018, and the Alibaba-AIR Program (22088682). We also thank the computational resource from the SJTU student innovation center.

\section*{Ethics Statement}
Our tense test set is based on the widely used public corpus Europarl in the field of machine translation. In creating this test set, we only corrected tense and description errors of some English references and did not change the original semantics, so there are no ethical issues arising.

\bibliography{anthology,custom}
\bibliographystyle{acl_natbib}

\appendix

\section{Online Translation}

\begin{itemize}

\item Google Translator: \url{https://translate.google.com/} as of December of 2022.

\item Bing Translator: \url{https://www.bing.com/translator} as of December of 2022.

\item DeepL Translator: \url{https://www.deepl.com/translator} as of December of 2022.

\end{itemize}

\section{Examples of Translators' Errors}
\label{sec:appendix}

Table \ref{tableB} shows several translating errors of common business translators. The display form is a group of five sentences: original French sentence, corresponding English reference, Bing translation, DeepL translation, and Google translation.

\begin{table*}[!htbp]\small
\centering
\begin{tabular}{p{0.65\linewidth} | p{0.30\linewidth}}
\hline
\textbf{Sentence} & \textbf{Tense}\\
\hline
\textbf{Origin: On avait fait des comparaisons.} &   \\
\textbf{Reference:} We {\color{orange}{had made}} comparisons. & \textit{Past perfect}  \\
\textbf{Bing:} Comparisons {\color{orange}{were made}}. &  \textit{{\color{blue}{Past simple}}} \\
\textbf{DeepL:} Comparisons {\color{orange}{were made}}.&  \textit{{\color{blue}{Past simple}}} \\
\textbf{Google:} We {\color{orange}{made}} comparisons.&  \textit{{\color{blue}{Past simple}}} \\
\hline
\textbf{Origin: Qui avait cru qu 'il serait facile de réunir l' Europe ?} &   \\
\textbf{Reference:} Who {\color{orange}{had thought}} that it {\color{orange}{would be}} easy to reunite Europe? & \textit{Past perfect+Modal}  \\
\textbf{Bing:} Who {\color{orange}{thought}} it {\color{orange}{would be}} easy to bring Europe together? &  \textit{{\color{blue}{Past simple}}+Modal} \\
\textbf{DeepL:} Who {\color{orange}{thought}} it {\color{orange}{would be}} easy to reunite Europe? &  \textit{{\color{blue}{Past simple}}+Modal} \\
\textbf{Google:} Who {\color{orange}{thought}} it {\color{orange}{would be}} easy to reunite Europe? &  \textit{{\color{blue}{Past simple}}+Modal} \\
\hline
\textbf{Origin: Je pensais avoir été assez clair.} &   \\
\textbf{Reference:} I {\color{orange}{thought}} I {\color{orange}{had been}} quite clear. & \textit{Past simple+Past perfect}  \\
\textbf{Bing:} I {\color{orange}{thought}} I {\color{orange}{was}} pretty clear. &  \textit{Past simple+{\color{blue}{Past simple}}} \\
\textbf{DeepL:} I {\color{orange}{thought}} I {\color{orange}{had made}} myself clear.&  \textit{Past simple+Past perfect} \\
\textbf{Google:} I {\color{orange}{thought}} I {\color{orange}{was}} clear enough.&  \textit{Past simple+{\color{blue}{Past simple}}} \\
\hline
\textbf{Origin: Un versement similaire avait eu lieu l 'année précédente.} &  \\
\textbf{Reference:} A similar payment {\color{orange}{had taken place}} in the previous year. & \textit{Past perfect}  \\
\textbf{Bing:} A similar payment {\color{orange}{had taken place}} the previous year. &  \textit{Past perfect} \\
\textbf{DeepL:} A similar payment {\color{orange}{was made}} the previous year.&  \textit{{\color{blue}{Past simple}}} \\
\textbf{Google:} A similar payment {\color{orange}{had taken place}} the previous year.	&  \textit{Past perfect} \\
\hline
\textbf{Origin: C 'est pour cela que la voie avait été tracée à Helsinki.} &   \\
\textbf{Reference:} That{\color{orange}{'s}} why the way {\color{orange}{had been paved}} in Helsinki.& \textit{Present simple+Past perfect}  \\
\textbf{Bing:} That {\color{orange}{is}} why the path {\color{orange}{was paved}} out in Helsinki. &  \textit{Present simple+{\color{blue}{Past simple}}} \\
\textbf{DeepL:} That {\color{orange}{is}} why the way {\color{orange}{was paved}} in Helsinki.&  \textit{Present simple+{\color{blue}{Past simple}}} \\
\textbf{Google:} This {\color{orange}{is}} why the way {\color{orange}{had been traced}} in Helsinki.&  \textit{Present simple+Past perfect} \\
\hline

\textbf{Origin: Je citerai pour exemple le vote à la majorité qualifiée.} &   \\
\textbf{Reference:} I {\color{orange}{will cite}} qualified majority voting as an example. & \textit{Future simple}  \\
\textbf{Bing:} One example {\color{orange}{is}} qualified majority voting. &  \textit{{\color{blue}{Present simple}}} \\
\textbf{DeepL:} An example {\color{orange}{is}} qualified majority voting. &  \textit{{\color{blue}{Present simple}}} \\
\textbf{Google:} I {\color{orange}{will cite}} as an example qualified majority voting. &  \textit{Future simple} \\
\hline
\textbf{Origin: Nous espérons tous qu 'elle finira.} &   \\
\textbf{Reference:} We all {\color{orange}{hope}} that it {\color{orange}{will come}} to an end. & \textit{Present simple+Future simple}  \\
\textbf{Bing:} We all {\color{orange}{hope}} that it {\color{orange}{will end}}. &  \textit{Present simple+Future simple} \\
\textbf{DeepL:} We all {\color{orange}{hope}} it {\color{orange}{will end}}. &  \textit{Present simple+Future simple}  \\
\textbf{Google:} We all {\color{orange}{hope}} it {\color{orange}{ends}}. &  \textit{Present simple+{\color{blue}{Present simple}}}  \\
\hline
\textbf{Origin: Que se passera-t-il si une nouvelle crise survient l 'année prochaine ?	} &   \\
\textbf{Reference:} What {\color{orange}{will happen}} if a new crisis {\color{orange}{occurs}} next year? & \textit{Future simple+Present simple}  \\
\textbf{Bing:} What {\color{orange}{will happen}} if a new crisis {\color{orange}{occurs}} next year? &  \textit{Future simple+Present simple}\\
\textbf{DeepL:} What {\color{orange}{happens}} if there {\color{orange}{is}} another crisis next year? &  \textit{{\color{blue}{Present simple}}+Present simple}  \\
\textbf{Google:} What {\color{orange}{will happen}} if a new crisis {\color{orange}{occurs}} next year? &  \textit{Future simple+Present simple}\\
\hline

\textbf{Origin: Nous en avons terminé avec les explications de vote.} &   \\
\textbf{Reference:} We {\color{orange}{have finished}} with the explanations of vote. & \textit{Present perfect}  \\
\textbf{Bing:} That {\color{orange}{concludes}} the explanations of vote. &  \textit{{\color{blue}{Present simple}}} \\
\textbf{DeepL:} That {\color{orange}{concludes}} the explanations of vote. &  \textit{{\color{blue}{Present simple}}} \\
\textbf{Google:} We {\color{orange}{have finished}} with the explanations of vote. &  \textit{Present perfect} \\
\hline
\textbf{Origin: Le fait est que le génie Internet est sorti de sa bouteille.} &   \\
\textbf{Reference:} The fact {\color{orange}{is}} that Internet genius {\color{orange}{has gone}} out of its bottle. & \textit{Present simple+Present perfect}  \\
\textbf{Bing:} The fact {\color{orange}{is}} that the Internet genie {\color{orange}{is}} out of the bottle. &  \textit{Present simple+{\color{blue}{Present simple}}} \\
\textbf{DeepL:} The fact {\color{orange}{is}} that the Internet genie {\color{orange}{is}} out of the bottle. &  \textit{Present simple+{\color{blue}{Present simple}}} \\
\textbf{Google:} The thing {\color{orange}{is}}, the internet genius {\color{orange}{is}} out of the bottle.	 &  \textit{Present simple+{\color{blue}{Present simple}}} \\
\hline
\textbf{Origin: Je voulais simplement le mentionner puisqu 'on a cité certains pays.} &   \\
\textbf{Reference:} I just {\color{orange}{wanted}} to mention that because some countries {\color{orange}{have been mentioned}}. & \textit{Past simple+Present perfect}  \\
\textbf{Bing:} I just {\color{orange}{wanted}} to mention this because some countries {\color{orange}{have been mentioned}}. &  \textit{Past simple+Present perfect} \\
\textbf{DeepL:} I just {\color{orange}{wanted}} to mention it because some countries {\color{orange}{were mentioned}}. &  \textit{Past simple+{\color{blue}{Past simple}}} \\
\textbf{Google:} I simply {\color{orange}{wanted}} to mention it since certain countries {\color{orange}{have been mentioned}}.	 &  \textit{Past simple+Present perfect} \\
\hline
\textbf{Origin: La dynamique de croissance et de création d 'emplois est évacuée.} &   \\
\textbf{Reference:} The dynamic of growth and job creation {\color{orange}{has run}} its course.	& \textit{Present perfect}  \\
\textbf{Bing:} The momentum for growth and job creation {\color{orange}{has been removed}}. &  \textit{Present perfect} \\
\textbf{DeepL:} The dynamics of growth and job creation {\color{orange}{are evacuated}}. &  \textit{{\color{blue}{Present simple}}} \\
\textbf{Google:} The dynamic of growth and job creation {\color{orange}{is evacuated}}. &  \textit{{\color{blue}{Present simple}}} \\
\hline

\end{tabular}
\caption{\label{tableB}
French-English utterances and corresponding translations by Bing, DeepL, Google translators. The {\color{orange}{words}}  in orange indicate the translated verbs. The tenses in {\color{blue}{blue}}  indicate the wrong predictions.}
\end{table*}

\section{Examples of Baseline Prediction Errors and Corresponding Annotations}
\label{sec:appendixC}

Table \ref{tableC} shows several examples of predictions and corresponding annotations of baselines in Section \ref{lab:exp}. Each group consists ten sentences, which are original French sentence, corresponding English reference, Transformer(tense-rich) prediction, Transformer(tense-poor) prediction, LSTM(tense-rich) prediction, LSTM(tense-poor) prediction, CNN(tense-rich) prediction, CNN(tense-poor) prediction, Bi-Transformer(tense-rich) prediction and Bi-Transformer(tense-poor) prediction.

\begin{table*}[!htbp]\small
\centering
\begin{tabular}{p{0.75\linewidth} | p{0.20\linewidth}}
\hline
\textbf{Sentence} & \textbf{Tense}\\
\hline
\textbf{Origin: J 'avais considéré que Mme Lulling était une Luxembourgeoise.} &   \\
\textbf{Reference:} I {\color{orange}{had assumed}}that Mrs Lulling {\color{orange}{was}} a Luxembourgoise.	& \textit{PasPerfect+Past}	  \\
\textbf{Transformer1:} I {\color{orange}{believed}} that Mrs Lulling {\color{orange}{was}} a Luxembourgois. &  \textit{{\color{blue}{Past}}+Past} \\
\textbf{Transformer2:} I {\color{orange}{considered}} that Mrs Lulling {\color{orange}{was}} a daughter. &  \textit{{\color{blue}{Past}}+Past} \\
\textbf{LSTM1:} I {\color{orange}{thought}} that Mrs Lulling {\color{orange}{was}} a Luxembourgoof. &  \textit{{\color{blue}{Past}}+Past}\\
\textbf{LSTM2:} I {\color{orange}{considered}} that Mrs Lulling {\color{orange}{was}} a stranglehold. &  \textit{{\color{blue}{Past}}+Past}\\
\textbf{CNN1:} I {\color{orange}{considered}} that Mrs Lulling {\color{orange}{was}} a Luxembourgo. &  \textit{{\color{blue}{Past}}+Past}\\
\textbf{CNN2:} In my view, Mrs Lulling {\color{orange}{was}} a Luxembourger. &  \textit{{\color{blue}{Past}}+Past} \\
\textbf{Bi-Transformer1:} I {\color{orange}{thought}} that Mrs Lulling {\color{orange}{was}} a Luxembourgois. &  \textit{{\color{blue}{Past}}+Past} \\
\textbf{Bi-Transformer2:} I {\color{orange}{thought}} that Mrs Lulling {\color{orange}{was}} a sort of Greens. &  \textit{{\color{blue}{Past}}+Past} \\
\hline
\textbf{Origin: Mais on les avait votés lors de la dernière période de session.} &   \\
\textbf{Reference:} However, they {\color{orange}{had been voted}} on at the last part-session.	& \textit{PasPerfect}	  \\
\textbf{Transformer1:} But we {\color{orange}{voted}} for them at the last part-session. &  \textit{{\color{blue}{Past}}} \\
\textbf{Transformer2:} But we {\color{orange}{voted}} for them at the last part-session. &  \textit{{\color{blue}{Past}}} \\
\textbf{LSTM1:} However, we {\color{orange}{had voted}} in favour of the last part-session. &  \textit{PasPerfect}\\
\textbf{LSTM2:} However, we {\color{orange}{had voted}} in the last part-session.&  \textit{PasPerfect}\\
\textbf{CNN1:} But we {\color{orange}{voted}} in the last part-session. &  \textit{{\color{blue}{Past}}}\\
\textbf{CNN2:} However, we {\color{orange}{voted}} in the last part-session.	 &  \textit{{\color{blue}{Past}}} \\
\textbf{Bi-Transformer1:} But we {\color{orange}{were voting}} on them at the last part-session. &  \textit{{\color{blue}{Past}}} \\
\textbf{Bi-Transformer2:} We, though, {\color{orange}{voted}} on them at the last part-session. &  \textit{{\color{blue}{Past}}} \\
\hline
\textbf{Origin: Il avait été averti par l 'association des employeurs irlandais.} &   \\
\textbf{Reference:} He {\color{orange}{had been alerted}} by the Irish employers' association.	& \textit{PasPerfect}	  \\
\textbf{Transformer1:} He {\color{orange}{was told}} it by the Irish employers' association. &  \textit{{\color{blue}{Past}}} \\
\textbf{Transformer2:} The Irish employers' association {\color{orange}{had warned.}} &  \textit{PasPerfect} \\
\textbf{LSTM1:} He {\color{orange}{was told}} it by the Irish employers' association. &  \textit{{\color{blue}{Past}}}\\
\textbf{LSTM2:} It {\color{orange}{was warned}} by the association of the Irish employers.&  \textit{{\color{blue}{Past}}}\\
\textbf{CNN1:} He {\color{orange}{was told}} by the Irish employers' association. &  \textit{{\color{blue}{Past}}}\\
\textbf{CNN2:} It {\color{orange}{was warned}} by the association of the Irish employers.	 &  \textit{{\color{blue}{Past}}} \\
\textbf{Bi-Transformer1:} He {\color{orange}{was told}} it by the Irish employers' association. &  \textit{{\color{blue}{Past}}} \\
\textbf{Bi-Transformer2:} The Irish employers' association {\color{orange}{had been notified}} by the Irish employers' association.	 &  \textit{PasPerfect} \\
\hline

\textbf{Origin: Je suis très curieux de voir ce que nous allons faire.} &   \\
\textbf{Reference:} I {\color{orange}{am}} very curious to see what we {\color{orange}{are}} going to do.		& \textit{Present}	  \\
\textbf{Transformer1:} I {\color{orange}{am}} very curious to see what we {\color{orange}{are}} going to do. &  \textit{Present} \\
\textbf{Transformer2:} I {\color{orange}{am}} very curious about what we {\color{orange}{are}} going to do. &  \textit{Present} \\
\textbf{LSTM1:} I {\color{orange}{am}} very curious to see what we {\color{orange}{will do}}. &  \textit{Present+{\color{blue}{Future}}}\\
\textbf{LSTM2:} I {\color{orange}{am}} very keen to see what we {\color{orange}{are}} going to do.&  \textit{Present}\\
\textbf{CNN1:} I {\color{orange}{am}} very curious to see what we {\color{orange}{are}} going to do. &  \textit{Present}\\
\textbf{CNN2:} I {\color{orange}{am}} very curious to see what we {\color{orange}{are}} going to do.	 &  \textit{Present} \\
\textbf{Bi-Transformer1:} I {\color{orange}{am}} very curious to see what we {\color{orange}{are}} going to do.&  \textit{Present} \\
\textbf{Bi-Transformer2:} I {\color{orange}{am}} very interested to see what we {\color{orange}{are}} going to do.	 &  \textit{Present} \\
\hline
\textbf{Origin: Nous espérons maintenant qu 'il va agir de façon énergique.} &   \\
\textbf{Reference:} We now {\color{orange}{hope}} that he {\color{orange}{is}} going to act decisively.	& \textit{Present}	  \\
\textbf{Transformer1:} We now {\color{orange}{hope}} that it {\color{orange}{will act}} decisively. & \textit{Present+{\color{blue}{Future}}}\\
\textbf{Transformer2:}{\color{orange}{Let}} us now {\color{orange}{hope}} that it {\color{orange}{will act}} energetically.&  \textit{Present+{\color{blue}{Future}}}\\
\textbf{LSTM1:} We now {\color{orange}{hope}} that it {\color{orange}{will}} act vigorously. &  \textit{Present+{\color{blue}{Future}}}\\
\textbf{LSTM2:} {\color{orange}{Let}} us {\color{orange}{hope}} now that it {\color{orange}{will act}} energetically.&  \textit{Present+{\color{blue}{Future}}}\\
\textbf{CNN1:} We now {\color{orange}{hope}} that it {\color{orange}{is}} going to act energetically. &  \textit{Present}\\
\textbf{CNN2:} {\color{orange}{Let}} us {\color{orange}{hope}} that it {\color{orange}{is}} going to act vigorously.	 &  \textit{Present} \\
\textbf{Bi-Transformer1:} We now {\color{orange}{hope}} that it {\color{orange}{will act}} vigorously.&  \textit{Present+{\color{blue}{Future}}} \\
\textbf{Bi-Transformer2:} {\color{orange}{Let}} us now {\color{orange}{hope}} that this {\color{orange}{will take}} a strong stand.	 &  \textit{Present+{\color{blue}{Future}}} \\
\hline

\textbf{Origin: D'ici là, je suis sûr que nous serons passés à au moins 27 États membres.} &   \\
\textbf{Reference:} By then, I {\color{orange}{am}} sure we {\color{orange}{will have enlarged}} to at least 27 Member States.	& \textit{Present+FutPerfect}	  \\
\textbf{Transformer1:} That {\color{orange}{is}} why I {\color{orange}{am}} sure that we {\color{orange}{will be left}} to at least 27 Member States. & \textit{Present+{\color{blue}{Future}}}\\
\textbf{Transformer2:} In this connection, I {\color{orange}{am}} sure we {\color{orange}{will have had}} at least 27 Member States.&  \textit{Present+FutPerfect}\\
\textbf{LSTM1:} I {\color{orange}{am}} sure that we {\color{orange}{will be}} at least 27 Member States. &  \textit{Present+{\color{blue}{Future}}}\\
\textbf{LSTM2:} That {\color{orange}{is}} why I {\color{orange}{am}} sure we {\color{orange}{will be}} at least 27 Member States.&  \textit{Present+{\color{blue}{Future}}}\\
\textbf{CNN1:} I {\color{orange}{am}} sure that we {\color{orange}{will be}} at least 27 Member States.& \textit{Present+{\color{blue}{Future}}}\\
\textbf{CNN2:} That {\color{orange}{is}} why I {\color{orange}{am}} sure we {\color{orange}{will be}} able to pass on at least 27 Member States. &  \textit{Present+{\color{blue}{Future}}} \\
\textbf{Bi-Transformer1:} I {\color{orange}{am}} sure that we {\color{orange}{will be}} doing so at least 27 Member States.&  \textit{Present+{\color{blue}{Future}}} \\
\textbf{Bi-Transformer2:} I {\color{orange}{am}} sure that we {\color{orange}{will have}} at least 27 Member States.	 &  \textit{Present+{\color{blue}{Future}}} \\
\hline

\end{tabular}
\caption{\label{tableC}
French-English utterances and corresponding predictions by baselines mentioned in Section \ref{lab:exp}. The {\color{orange}{words}}  in orange indicate the translated verbs. The tenses in {\color{blue}{blue}}  indicate the wrong predictions. }
\end{table*}

\section{Additional Notes On Human Review}
\label{sec:human}
\subsection{Recruitment of Human Reviewers}
We recruited reviewers from students majoring in French. Taking Diplôme Approfondi de Langue Française(DALF) C1 French exam results, International English Language Testing System(IELTS) exam results, and their GPA in French courses into account, we recruited 2 reviewers in the same country of the authors' at last.

\subsection{Instructions Given to Reviewers}
We offer the annotation rules in Section \ref{sec2}, and require the reviewers to accomplish the following tasks：

\begin{itemize}
    \item Determine whether the tense of the English translation is accurate and reasonable. If not, give an English translation that you consider reasonable.
    \item Determine whether the meaning of the English translation is correct. If not, give an English translation that you consider reasonable.
    \item Determine whether the corresponding tense label of the English translation is correct according to the natural language understanding.
\end{itemize}

\section{Experimental Setup}
\subsection{Model}
Table~\ref{tab:model} provides the number of parameters, training budget, and hyperparameters of each model.
All experiments were performed on a single V100 GPU and the hyperparameters are by default. We report the result of a single run for each experiment.

\begin{table*}[htpb]
    \centering
    \begin{tabular}{l | c c | c c}
    \toprule
       \multirow{2}{*}{\bf Model}  &  \multirow{2}{*}{\bf \# Param.} &  \multirow{2}{*}{\bf GPU Hours} & \multicolumn{2}{c}{\bf Hyperparam.}\\ 
       & & & learning rate & dropout \\
       \midrule
    Transformer   & 83M & 0.9h & 5e-4 & 0.3\\
    LSTM   & 58M & 0.8h & 1e-3 & 0.2\\   
    CNN   & 30M &  0.7h & 0.25 & 0.2\\
    Bi-Transformer & 83M & 1.7h & 5e-4 & 0.3\\
    \bottomrule
    \end{tabular}
    \caption{The number of parameters, training budget (in GPU hours), and hyperparameters of each model.}
    \label{tab:model}
\end{table*}

\subsection{Data}
Table~\ref{tab:data} shows the data statistics we used in this paper.
\begin{table*}[htpb]
    \centering
    \begin{tabular}{l l r l }
    \toprule
        \bf Split & \bf Name & \bf \# Sent. & \bf Domain \\
        \midrule
         \multirow{2}{*}{\bf Train} & Train set from EuroparlTR (tense-rich) & 97K & Politics \\
                                    & Train set from Europarl (tense-poor) & 97K & Politics\\
                                    \midrule
         \multirow{3}{*}{\bf Test}  & Tense set & 552 & Politics\\
                                    & Europarl test set & 2950 & Politics\\   
                                    & WMT15 test set & 3003 & News\\    
                                     \midrule
         \multirow{1}{*}{\bf Valid}  & Valid set from EuroparlTR (tense-rich) & 717 & Politics \\
    \bottomrule
    \end{tabular}
    \caption{Data statistics. Training data has been filtered to avoid data leakage.}
    \label{tab:data}
\end{table*}

\subsection{Packages}
Table~\ref{tab:package} shows the packages we used for preprocessing, model training, evaluation and tense labeling.

\begin{table*}[htpb]
    \centering
    \begin{threeparttable}
        \begin{tabular}{l l l}
        \toprule
             \bf Usage & \bf Package & \bf License  \\
             \midrule
             \multirow{2}{*}{Preprocessing} & mosesdecoder~\cite{koehn-etal-2007-moses}\tnote{1} &  LGPL-2.1 \\
             & subword-nmt~\cite{sennrich-etal-2016-neural}\tnote{2} & MIT \\
             \midrule
             Model training & fairseq~\cite{ott2019fairseq}\tnote{3} & MIT \\
             \midrule
             \multirow{2}{*}{Evaluation} &  SacreBLEU~\cite{post2018call}\tnote{4}& Apache 2.0 \\
             & COMET~\cite{rei2020comet}\tnote{5} & Apache 2.0 \\
             \midrule
             Tense labeling & spaCy~\cite{Honnibal_spaCy_Industrial-strength_Natural_2020}\tnote{6} & MIT \\
             \bottomrule
        
        \end{tabular}
        \begin{tablenotes}
            \item[1] \url{https://github.com/moses-smt/mosesdecoder}\\
            \item[2] \url{https://github.com/rsennrich/subword-nmt} \\
            \item[3] \url{https://github.com/facebookresearch/fairseq} \\
            \item[4] \url{https://github.com/mjpost/sacrebleu}\\
            (nrefs:1|case:mixed|eff:no|tok:13a|smooth:exp|version:2.3.1)  \\
            \item[5] \url{https://github.com/Unbabel/COMET}\\
            (wmt20-comet-da) \\
            \item[6] \url{https://github.com/explosion/spaCy}
        \end{tablenotes}
    \end{threeparttable}
    \caption{Packages we used for preprocessing, model training, evaluation and tense labeling.}
    \label{tab:package}
\end{table*}

\end{CJK*}
\end{document}